\documentclass[10pt,twocolumn,letterpaper]{article}

 \usepackage[pagenumbers]{cvpr} % To force page numbers, e.g. for an arXiv version

\usepackage{amsmath}
\newcommand{\m}[1]{\mathcal{#1}}
\DeclareMathOperator*{\argmin}{arg\,min}
\usepackage[dvipsnames]{xcolor}

\usepackage{xspace}
\newcommand{\agoodname}{\textit{Holo-Relighting}\xspace}

\definecolor{cvprblue}{rgb}{0.21,0.49,0.74}
\usepackage[pagebackref,breaklinks,colorlinks,citecolor=cvprblue]{hyperref}

\usepackage{comment}

\makeatletter
\newcommand{\printfnsymbol}[1]{%
	\textsuperscript{\@fnsymbol{#1}}%
}
\makeatother
\usepackage{footnote}
\usepackage[symbol]{footmisc}

\def\paperID{2179} % *** Enter the Paper ID here
\def\confName{CVPR}
\def\confYear{2024}
\begin{document}
\vspace{-10mm}
\title{{Holo-Relighting: Controllable Volumetric Portrait Relighting \\ from a Single Image}}

%%%%%%%%% AUTHORS - PLEASE UPDATE
\author{%
  Yiqun~Mei$^{1}$\quad Yu Zeng$^{1}$\printfnsymbol{2}\quad He Zhang$^{2}$\printfnsymbol{2} \quad Zhixin Shu$^{2}$\printfnsymbol{2}\quad Xuaner Zhang$^{2}$\quad Sai Bi$^{2}$\\ Jianming Zhang$^{2}$\quad HyunJoon Jung$^{2}$\quad Vishal M.~Patel$^{1}$ \\
  {\small$^{1}$Johns Hopkins University\quad \quad $^{2}$Adobe Inc.}
}
\twocolumn[{%
\renewcommand\twocolumn[1][]{#1}%
\maketitle
\vspace{-9mm}
\begin{center}
    \centering
    \includegraphics[trim=0cm 0.9cm 0cm 0cm, width=1\textwidth]{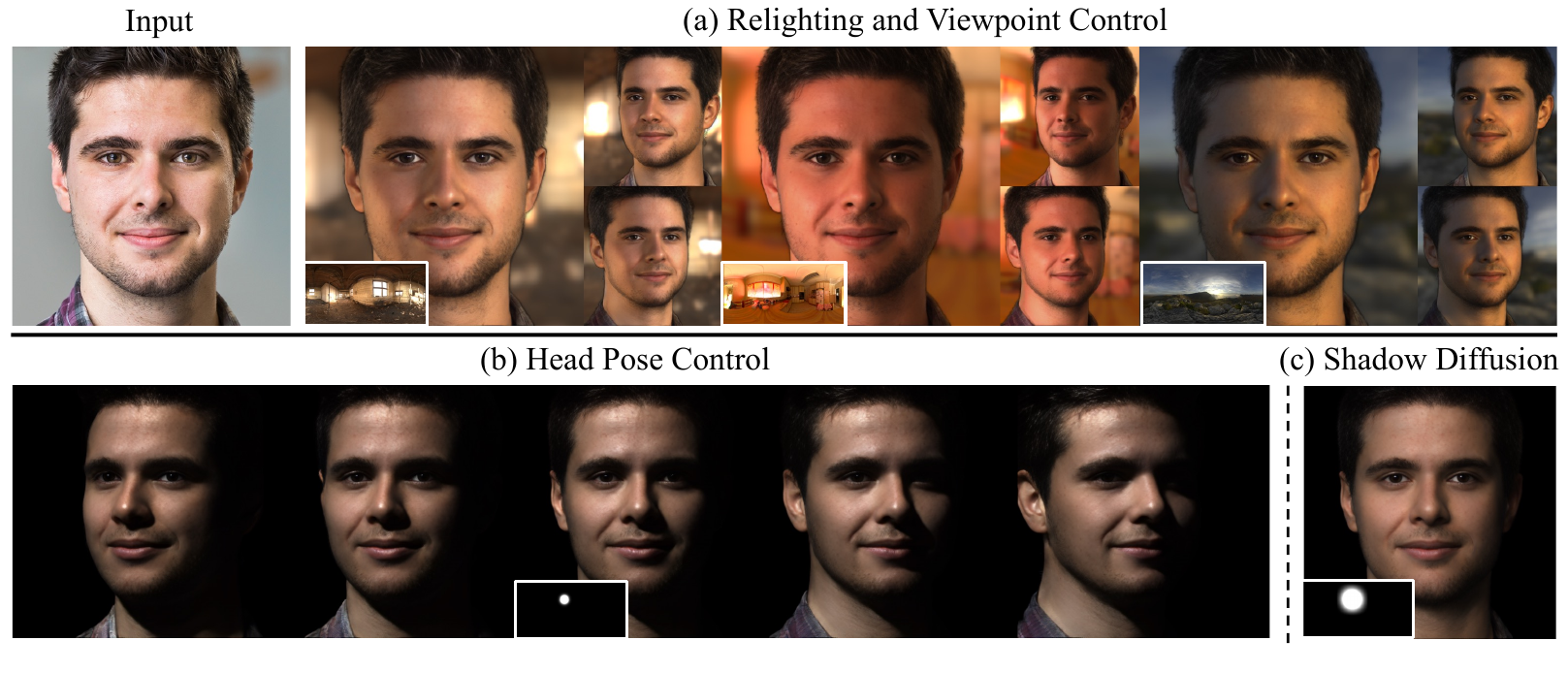}

    \captionof{figure}{
 \agoodname performs volumetric relighting on a single input portrait image, allowing users to individually control (1) lighting effect via an environment map, (2) camera viewpoint and (3) head pose. It is highly expressive and can render complex illumination effects on in-the-wild human faces with accurate view consistency (a
 ). Controls are well disentangled to produce a realistic rendering of moving shadows cast by a point light while rotating the head (b). Practical photographic applications such as shadow diffusion (softening) are also made feasible with our system (c).
    }
   \label{fig:teaser} 
\end{center}%
}]

\footnotetext[2]{The second authors}
\renewcommand*{\thefootnote}{\fnsymbol{footnote}}
\setcounter{footnote}{0}

\begin{abstract}
At the core of portrait photography is the search for ideal lighting and viewpoint. The process often requires advanced knowledge in photography and an elaborate studio setup. In this work, we propose \textit{Holo-Relighting}, a volumetric relighting method that is capable of synthesizing novel viewpoints, \textbf{and} novel lighting from a single image. \textit{Holo-Relighting} leverages the pretrained 3D GAN (EG3D) to reconstruct geometry and appearance from an input portrait as a set of 3D-aware features. We design a relighting module conditioned on a given lighting to process these features, and predict a relit 3D representation in the form of a tri-plane, which can render to an arbitrary viewpoint through volume rendering. Besides viewpoint and lighting control, \textit{Holo-Relighting} also takes the head pose as a condition to enable head-pose-dependent lighting effects. 
% , an under-explored relighting feature.
With these novel designs, \textit{Holo-Relighting} can generate complex non-Lambertian lighting effects (e.g., specular highlights and cast shadows) without using any explicit physical lighting priors. We train \textit{Holo-Relighting} with data captured with a light stage, and propose two data-rendering techniques to improve the data quality for training the volumetric relighting system. Through quantitative and qualitative experiments, we demonstrate \textit{Holo-Relighting} can achieve state-of-the-arts relighting quality with better photorealism, 3D consistency and controllability.

\end{abstract}

\section{Introduction}
%The essential goal and 
One essential challenge in portrait photography is to find the ideal lighting condition and viewpoint that best portray the subject, a process that often involves tedious adjustments of camera and lighting setup in a professional studio environment with expensive equipment~\cite{grey2014master,schriever1909complete}. These requirements, however, are beyond the reach of most consumer photographers, especially with the increasing demands of quick selfies and candid photos.
In this paper, we aim to meet the growing need of virtually synthesizing novel views and novel lighting for a given portrait to enable flexible portrait editing.

Several recent works~\cite{pandey2021total, shu2017portrait, sun2019single, zhou2019deep,Hou_2022_CVPR,zhang2020portrait,wang2020single,yeh2022learning,Hou_2021_CVPR, mei2023lightpainter, nvpr} have made progress on single-image portrait relighting that enables post-capture lighting editing. However, simultaneous relighting and view synthesis for headshot portrait have received less attention. Such task inherently requires 3D understanding of an image, and is often addressed under a multi-view setting with specialized acquisition mechanisms~\cite{guo2019relightables, zhang2021neural, meka2020deep, sun2021nelf, ma2007rapid,bi2021deep}. Despite great performance, these works are less approachable for average users and tend to fail on in-the-wild images, as they are designed for more controlled settings.

Precise view and lighting control requires a good estimation of physical properties such as material and geometry, which are fundamentally difficult given only a monocular 2D image. Previous works~\cite{deng2023lumigan,ranjan2023facelit,tan2022volux, pan2021shading, jiang2023nerffacelighting} usually rely on explicit physical modeling with simplified assumptions on the reflectance or/and lighting model to achieve the desired view and lighting control.
For example, Lambertian and Phong models are often used as simplified reflectance models~\cite{pan2021shading,ranjan2023facelit, tan2022volux}; spherical harmonics~\cite{deng2023lumigan, jiang2023nerffacelighting, ranjan2023facelit} is commonly used as a lighting representation. Other works~\cite{jiang2023nerffacelighting} make assumptions on the color of illumination to simplify their optimization space.

While these assumptions provide a reasonable approximation to the actual light transport, they suffer from limited expressiveness and result in producing unrealistic shading effects and less accurate lighting effects.

In this paper, we propose \textbf{\agoodname}, a controllable volumetric relighting method that can render novel views and novel lighting from a single image. We show that challenging high-frequency lighting effects can be handled in a fully implicit manner without relying on any physical reflectance and lighting models. The key idea of our method is to represent 3D information from the input image as a set of 3D-aware features by exploiting the inversion of a pretrained 3D GAN (EG3D~\cite{chan2022efficient}). We are then able to train a relighting model that is conditioned on the target lighting using these 3D-aware features. The output of the relighting model is a 3D representation embedded with the target illumination in the form of a tri-plane, from which an arbitrary view can be rendered using volume rendering. Besides viewpoint and lighting control, \agoodname also takes the head pose as an input condition and enables head-pose-dependent lighting effects, which is under-explored in prior works.
% Such user control has not been explored in existing methods.

We train the relighting module on a diverse set of portrait images under different illumination conditions, which are renderings using the ``one-light-at-a-time" (OLAT)
data captured with a light stage~\cite{debevec2000acquiring}.
Our goal is to train the relighting module to learn arbitrary lighting effects, including ones that demand accurate geometry such as cast shadow.
Therefore, it is crucial to ensure that the inverted input latent code retains more accurate geometry information.

Simply relying on existing GAN inversion techniques~\cite{roich2022pivotal,richardson2021encoding,xie2023high, yuan2023make,yin20233d} is sub-optimal due to the depth ambiguity when inferring geometry from a monocular input. To address this challenge, we extend the existing GAN inversion method~\cite{xie2023high} with multi-view regularization and camera pose optimization to encode a more accurate geometry. We additionally propose a shading transfer technique inspired by quotient image~\cite{shashua2001quotient} to accommodate for the misalignment in high-frequency details induced by imperfect inversion. We show that training with data created using these strategies is important for synthesizing high-quality shading effects.

With extensive quantitative and qualitative experiments,
we show that \agoodname can produce relit portraits with more flexible control, superior photo-realism and multi-view consistency compared to the state-of-the-art alternatives. We summarize our contributions as follows: 
\begin{itemize}[noitemsep]
\item A novel controllable volumetric relighting method that renders free-view relit portraits with state-of-the-art photo-realism and view consistency.
\item A novel relighting module capable of rendering complex shading effects including non-Lambertian reflections and cast shadows without imposing physical constraints.

\item Two data-rendering techniques to make more effective use of light stage captures for training a volumetric relighting system.

\end{itemize}

\section{Related Work}
\noindent\textbf{Portrait Relighting.} There is a significant body of work studying 2D portrait relighting over past decades. In the pioneering work, Debevec \etal~\cite{debevec2000acquiring} design a special capture rig \ie light stage to record reflectance field, which is then used to render images with novel lighting. Later methods remove hardware requirement by using techniques such as style transfer~\cite{shu2017portrait, shih2014style}, quotient image~\cite{shashua2001quotient,peers2007post}, and intrinsic decomposition~\cite{Hou_2022_CVPR,ji2022relight,le2019illumination,barron2014shape, shahlaei2015realistic, li2014intrinsic}. Several recent approaches~\cite{nestmeyer2020learning,pandey2021total,sun2019single,zhou2019deep,zhang2021neural,nvpr, mei2023lightpainter, wang2020single, yeh2022learning} use neural networks to synthesize lighting effects and achieve higher photorealism. Our method is motivated by their success.  

\begin{figure*}[t!]
	\centering
	\includegraphics[clip, trim=0cm 8.5cm 2.7cm 0cm, width=0.98\textwidth]{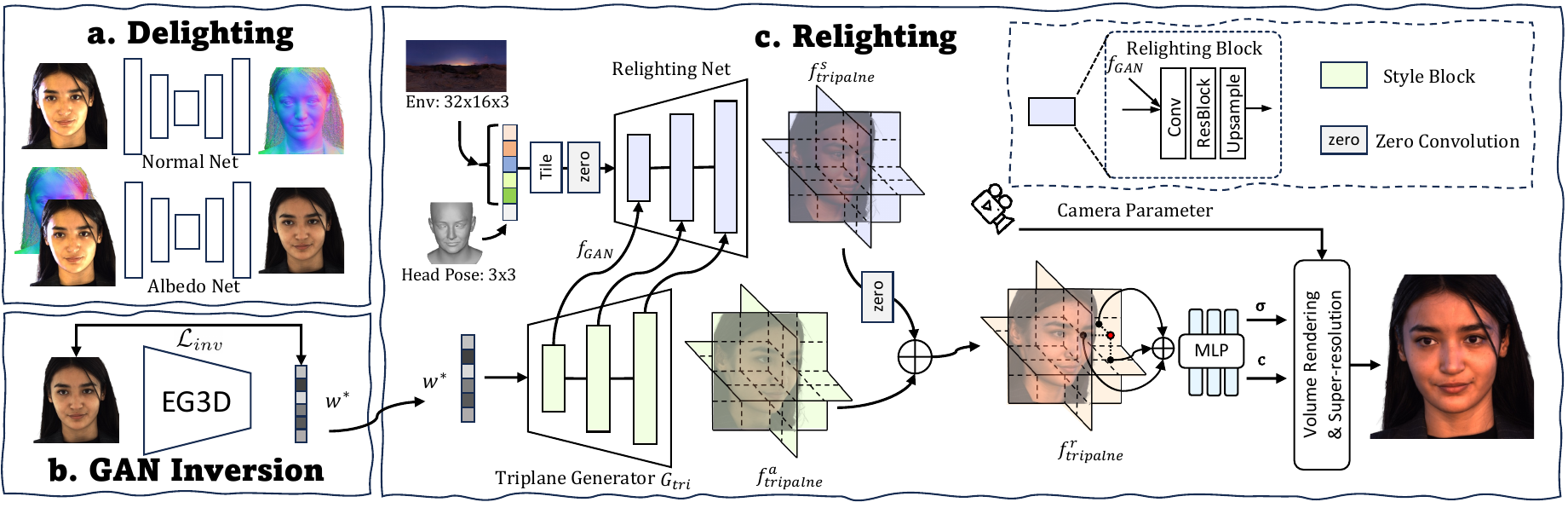}
\vskip-16pt	\caption{\textbf{An overview of \agoodname.} Our method consists of three stages. (a) We first remove the shading from the input portrait and estimate an albedo image. (b) We then conduct GAN inversion upon EG3D to obtain a latent code $w^{*}$ encoding 3D information of the subject. (c) The relighting network takes in the lighting condition, head pose as well as intermediate GAN features produced by EG3D's tri-plane generator $G_{tri}$ using the inverted latent code $w^{*}$, and predicts a shading tri-plane $f^{s}_{triplane}$, which is summed to the albedo tri-plane $f^{a}_{triplane}$, resulting in the relit tri-plane $f^{r}_{triplane}$ with lighting embedded. High-resolution RGB images can be rendered from $f^{r}_{triplane}$ via volume rendering and a super-resolution network. During training, we freeze $G_{tri}$ and only update the relighting net.}\label{workflow}
\vspace{-6mm}
\end{figure*}
Beyond a 2D portrait, very few efforts investigate relighting and view synthesis in a unified framework. Many of them rely on multi-view setup to construct a physical relighting model~\cite{guo2019relightables, ma2007rapid,bi2021deep,wang2023sunstage} or interpolate light transport~\cite{zhang2021neural, meka2020deep, sun2021nelf, prao2022vorf}, thus are not suitable for in-the-wild images. For a single 2D input, the problem remains challenging due to its ill-posed nature. Some attempts use handcrafted face priors such as 3DMM~\cite{blanz2023morphable} and apply physical or neural rendering ~\cite{Paraperas_2023_ICCV, wang2023free,tewari2021monocular} for relighting. These methods often lack geometric details and fail to synthesize hairs and mouth interior. As a result, their realism lags behind existing 2D approaches.

\vspace{-4mm}
\paragraph{3D GANs.} Recent 3D Generative Adversarial Networks (GANs)~\cite{goodfellow2020generative} can generate view-consistent images by training on 2D images.  Early 3D GANs embed explicit 3D structure such as voxels \cite{henzler2019escaping,nguyen2019hologan}, and meshes~\cite{chen2019learning,henderson2020leveraging,kanazawa2018learning} into their network but fail to achieve similar quality as those 2D GANs. More recent 3D GANs~\cite{chan2022efficient,or2022stylesdf,gu2021stylenerf, deng2022gram,schwarz2020graf,niemeyer2021giraffe,zhou2021cips, chan2021pi} adopt implicit 3D representation such as radiance field~\cite{mildenhall2021nerf}. Among them, EG3D~\cite{chan2022efficient} introduces an efficient tri-plane-based 3D representation which offers efficient and high-quality view synthesis on par with existing 2D GANs~\cite{karras2020analyzing,brock2018large}. Our method also adopts a tri-plane-based 3D representation for volume rendering. 

Most 3D GANs are not relightable. To enable lighting control, recent efforts propose to embed physical reflectance models, such as Lambertian~\cite{pan2021shading}, Phong~\cite{tan2022volux, ranjan2023facelit}, or Radiance Transfer~\cite{deng2023lumigan}, into their models to explicitly model lighting. While the simplified reflectance and illumination model (\ie Spherical Harmonics lighting~\cite{green2003spherical}) provide a good approximation to the light transport, they suffer from limited expressiveness and result in unrealistic rendering quality. Recently, Jiang \etal~\cite{jiang2023nerffacelighting} propose to distill shading from 3D GANs. But their method assumes a fixed illuminant color to simplify their optimization space, thus is not applicable for general-purpose relighting.
\vspace{-4mm}
\paragraph{GAN Inversion.} GAN inversion is an essential step for GAN-based image editing which maps an image back to the latent space of a pretrained generator. It has numerous applications including inpainting~\cite{richardson2021encoding,daras2021intermediate}, style transfer~\cite{yang2022vtoonify,yang2022pastiche} and restoration~\cite{wang2021towards,menon2020pulse, yang2021gan}. 2D GAN inversion also enables viewpoint and lighting control~\cite{chandran2021rendering,tewari2020stylerig,tewari2020pie,br2021photoapp}, but often lack view-consistency and/or limit to simple SH illumination. Recent efforts extend GAN inversion for 3D GANs to enable 3D face editing~\cite{xie2023high,yin20233d, yuan2023make}. Our method exploits 3D GAN inversion~\cite{xia2022gan} to retrieve 3D representations from an input image to help volumetric relighting. 
\vspace{-5mm}
\paragraph{Preliminary: EG3D Framework.}
Our method adopts pretrained EG3D~\cite{chan2022efficient} to extract
3D information from the input image through inversion. Here we briefly describe its framework. The core design of EG3D is to use an efficient tri-plane representation for volume rendering. A tri-plane consists of three individual 2D feature maps, where each of them represents three orthogonal plane ${f_{XY}, f_{YZ}, f_{XZ}}$ in 3D space. These feature maps are split channel-wise from a single feature map produced by its StyleGAN2-like~\cite{karras2020analyzing} tri-plane generator. 
For volume rendering, sampled 3D points along a ray are projected onto each plane to retrieve a summed 1-D feature $f$, from which a color feature $c$ and density $\sigma$ are decoded by an MLP. A multi-channel feature image $I_{c}$ is then obtained via volume rendering~\cite{mildenhall2021nerf}:
 \setlength{\belowdisplayskip}{2pt} \setlength{\belowdisplayshortskip}{2pt}
\setlength{\abovedisplayskip}{2pt} \setlength{\abovedisplayshortskip}{2pt}
\begin{align}
I_{c}(r) = \int_{t_n}^{t_f} T(t) \cdot \sigma(t) \cdot c(t) \, dt
\end{align}
where $T(t) = \exp\left(-\int_{t_n}^{t} \sigma(s) \, ds\right)$ is the transmittance. For efficiency, $I_{c}$ is accumulated at low resolution and up-sampled to the final high-resolution image through a super-resolution network.

\section{Method}

\agoodname takes a single portrait image as input and predicts a novel image under desired lighting condition, viewpoint and head pose. Specifically, our system consists of three stages (Figure~\ref{workflow}):  (1) Delighting: given an input image, we remove the shading and shadow effects and predict an albedo image (Figure~\ref{workflow}-(a));  (2) GAN Inversion: we reconstruct 3D information of the albedo image through GAN inversion, by projecting it to the latent space of a pretrained EG3D~\cite{chan2022efficient} (Figure~\ref{workflow}-(b)). The obtained latent code $w^{*}$ is used to retrieve a set of features $f_{GAN}$ from EG3D's tri-plane Generator $G$ and transmit them to a relighting network; (3) Relighting: we use a relighting module (Figure~\ref{workflow}-(c)) conditioned on a given environment map, head pose and camera pose to process the GAN features $f_{GAN}$ to produce a relit tri-plane $f_{triplane}^{r}$, from which a novel image can be synthesized through volume rendering.

\subsection{Delighting stage } \label{delight}

Following the popular delight-then-relight scheme in recent 2D relighting methods~\cite{mei2023lightpainter, pandey2021total, wang2020single, yeh2022learning}, we start from a delighting stage, which predicts an albedo image using two separate networks,~\ie the \textit{normal net} conditioned on the original input and the \textit{albedo net} conditioned on the both inferred normal and input, as shown in Figure~\ref{workflow}-(a). We use a U-Net architecture~\cite{ronneberger2015u} for both networks. Training details can be found in Section~\ref{data}.

\subsection{GAN Inversion stage } \label{inv}

We lift a 2D albedo image into 3D space by projecting it into the latent space of EG3D~\cite{chan2022efficient} through GAN inversion, which reconstructs 3D information of the subject. Specifically, given an albedo image $\m{A}$ and a pretrained EG3D denoted as $G$ and parameterized by $\theta$, GAN inversion stage involves searching for a latent code $w^{*}$ and fine-tuning $G$ to best reconstruct $\m{A}$. Formally, 
\begin{align}
    w^{*}, \theta^{*} = \argmin_{w, \theta} \mathcal{L}_{inv}(G(w;\theta), \m{A})
\end{align} where $\mathcal{L}_{inv}$ is the reconstruction loss. And $G(w;\theta)$ is the generated albedo using weights $\theta$. During optimization, we only update EG3D's tri-plane generator (denoted as $G_{tri}$) and freeze other parts. For simplicity, we refer $\theta$ as the parameters belonging to $G_{tri}$ in the following paragraphs.

With optimized $w^{*}$ and $\theta^{*}$, we can retrieve a set of intermediate features ${f_{GAN}}$ and an albedo tri-plane $f^{a}_{tripane}$ from $G_{tri}(w^{*};\theta^{*})$, which encodes the 3D information of the albedo image $\m{A}$. 
We adapt an existing inversion method~\cite{xie2023high} during inference and refer readers to their paper for more detail. When creating training data, we propose two additional strategies to align encoded geometry and appearance with the target image. Details are described in Section~\ref{training-data}.

\subsection{Relighting stage } \label{relight}
One core design of \agoodname is the use of a learned neural network for relighting, that can generate complex non-Lambertian reflections and cast shadows without using any physical lighting priors. In the following, we first describe the designed relighting network in detail and then introduce two data-rendering techniques to facilitate training.

\subsubsection{Relighting Net}

As shown in Figure~\ref{workflow}-(c), the relighting network takes a given lighting condition (\ie environment map) and head pose ($3\times 3$ rotation matrix) as input. They are first reshaped into 1D tensors, and then tiled to 2D maps before feeding into the relighting net. The relighting net adopts a pyramidal structure which progressively increases its spatial resolution to match that of the tri-plane generator $G_{tri}$. At each resolution stage, the relighting net takes in an intermediate GAN feature $f_{GAN}$ produced by $G_{tri}(w^{*};\theta^{*})$ of same resolution through concatenation, and outputs an upsampled feature for next resolution stage. The relighting net final produces a shading tri-plane $f^{s}_{triplane}$ which is added back to the albedo tri-plane $f^{a}_{triplane}$ to produce a relit tri-plane $f^{r}_{triplane}$
with the target illumination embedded. A relit image can then be rendered with volume rendering. In the following, we discuss several important aspects of the relighting net. 
\\

\vspace{-1mm}
\noindent\textbf{Feature-based Coarse-to-Fine Relighting.} We build the relighting net upon GAN features, which encode rich 3D properties of the subject. Using $f_{GAN}$, the network managed to render complex lighting effects such as specular highlights and cast shadows (Figure~\ref{fig:teaser}) that require a good understanding of the geometry. The relighting network is naturally designed to consume GAN features in a coarse-to-fine manner, which enables to synthesize a global lighting distribution first and then refine local details based on a global context. As such, the network learns to synthesize both diffuse and high-frequency specular highlights well.

\vspace{-3mm}
\paragraph{Stabilize Training with Zero-Convolution.} 
As we add the shading tri-plane $f^{s}_{triplane}$ back to the albedo tri-plane $f^{a}_{triplane}$, 
the relighting network training is prone to diverge at the beginning. Inspired by~\cite{zhang2023adding}, we apply two zero-initialized convolutions to the relighting network: one before the addition of shading tri-plane $f^{s}_{triplane}$ and albedo tri-plane $f^{a}_{triplane}$ and the other before taking in the lighting and pose condition.  This strategy allows the model to gradually incorporate the given illumination signals and thus stabilize training. 

\vspace{-3mm}
\paragraph{Model Head-Pose Dependent Shading.} Given a lighting condition, our relighting network can generate head-pose dependent effects,~\eg the moving specular highlights and cast shadows with changing head pose. This effect is often neglected in existing approaches~\cite{pan2021shading,deng2023lumigan,ranjan2023facelit,tan2022volux}. Although existing free-view relighting methods can render different views of a portrait, their view changes are always based on the movement of camera rather than head-pose.\footnote{Please refer to Fig.~\ref{fig:shadow_syth}~(b) and (c) for the difference between the shading change caused by viewpoint change and head-pose change.} In contrast, our method explicitly takes the head pose as an input condition, and learns to generate head-pose dependent shading/shadows through training.

\begin{figure}[t!]
	\centering
	\includegraphics[clip, trim=0.8cm 9.3cm 16.6cm 0cm, width=0.43\textwidth]{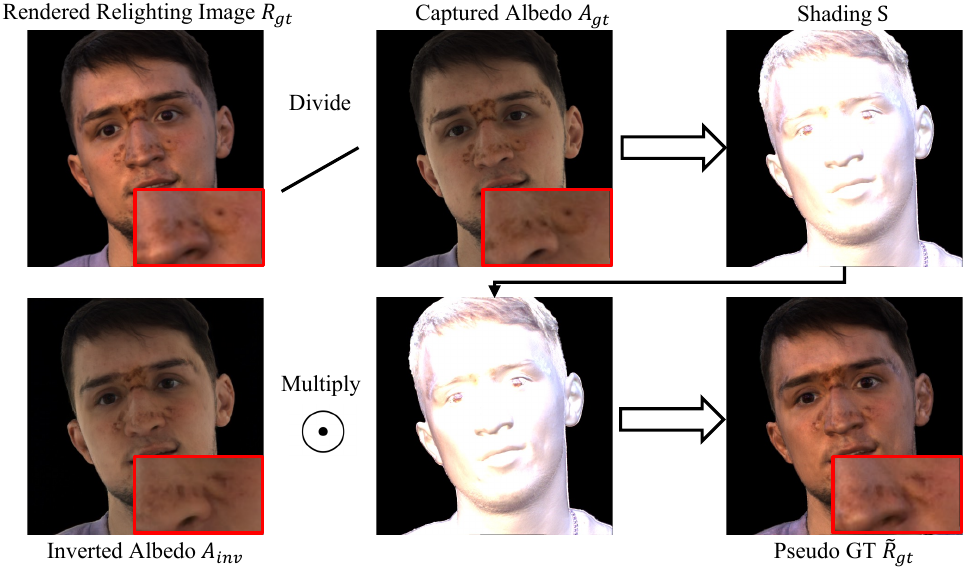}
\vskip-5pt	\caption{Illustration of proposed portrait shading transfer. We create a pseudo ground-truth image $\Tilde{\m{R}}_{gt}$ by transferring the shading from the OLAT renders. $\Tilde{\m{R}}_{gt}$ contains target illumination, and is consistent with the appearance encoded in $f_{GAN}$ (\ie $\m{A}_{inv}$).}\label{fig:shading_transfer}
\vspace{-7mm}
\end{figure}

\subsection{Training Data Creation}
\label{training-data}
Similar to prior works~\cite{mei2023lightpainter,pandey2021total,sun2019single}, we train the relighting net using light stage~\cite{debevec2000acquiring} data rendered with diverse lighting environments.
During training, the target image is a rendered relit image, and the input to the relighting net contains the target environment map, estimated head pose, camera pose, and GAN features $f_{GAN}$ from $G(w^{*};\theta^{*})$. 
As we aim to use the relighting model to learn geometry-dependent lighting effects, we need to ensure that $f_{GAN}$ encode accurate geometry of the input subject. We introduce a multi-view inversion strategy to leverage the multi-view light stage data for better geometry encoding. We then propose a shading transfer technique to deal with
misalignment in high-frequency details induced by imperfect inversion.

\vspace{-5mm}
\paragraph{Multi-View Regularization.} Single-view inversion inherently suffers from depth ambiguity. The incorrectly encoded geometry prevents the relighting net from using 3D-aware features for relighting. To address this issue, we extend~\cite{xie2023high} into a multi-view scheme by leveraging our multi-view light stage captures.
Formally, given a set of albedo images $\mathbf{\m{A}} = \{\m{A}^{1},\m{A}^{2},...,\m{A}^{n}\}$ captured from N viewpoints and associated camera poses $\mathbf{P}=\{p^{1},p^{2}...,p^{n}\}$, we first freeze $G$ and search for an optimal $w^{*}$ shared among the frames and per-frame pose $\mathbf{P}^{*}$ by minimizing:
\begin{align}
    \mathcal{L}_{w, \mathbf{P}} = \frac{1}{n}\sum_{1}^{n}\|\phi_{VGG}(G(w,p^{i})) -\phi_{VGG}(\m{A}^{i})\|_{2}
\end{align}
where $\phi_{VGG}$ is a pretrained VGG~\cite{simonyan2014very} feature extractor. In the second step, we freeze the derived $w^{*}$ and $\mathbf{P}^{*}$ and finetune the generator's parameter $\theta$ by optimizing
\begin{align}
  \mathcal{L}_{\theta} = \frac{1}{n}\sum_{1}^{n} \|G(\theta) - \m{A}^{i}\|_{2} +
  \mathcal{L}_{lpips}(G(\theta), \m{A}^{i})
\end{align}
where $\mathcal{L}_{lpips}$ is the perceptual loss defined in~\cite{zhang2018unreasonable}. With multi-view regularization, the retrieved GAN features can reflect more accurate geometry.

\vspace{-3mm}
\paragraph{Portrait Shading Transfer.} As shown in Figure \ref{fig:shading_transfer}, even with advanced inversion techniques, the inverted albedo (which is encoded in $f_{GAN}$) cannot perfectly capture the details from the input. When training with per-pixel loss, such texture differences lead to blurry results. Inspired by quotient image~\cite{shashua2001quotient}, we resolve this with a simple yet effective strategy,~\ie by transferring the lighting from the target OLAT rendering to the inverted albedo image to create a pseudo ground-truth relighting image. Formally, given the real ground-truth relighting image $\m{R}_{gt}$ and its associated albedo $\m{A}_{gt}$, we compute a shading image $\m{S} = \m{R}_{gt}/\m{A}_{gt}$ via element-wise division, a decomposition similar to~\cite{ichim2015dynamic}. We then transfer the shading $\m{S}$ to the albedo image $\m{A}_{inv}$ reconstructed from the latent code with a per-pixel multiplication $\m{S}$ \ie $\m{\Tilde{R}}_{gt} = \m{S} \odot \m{A}_{inv}$, where $\odot$ denotes the Hammond product, to produce a pseudo ground-truth image $\m{\Tilde{R}}_{gt}$, and use it to supervise the relighting module. As shown in Figure~\ref{fig:shading_transfer}, $\m{\Tilde{R}}_{gt}$ preserves target shading while containing consistent facial details with the inverted albedo image, and thus is also consistent with encoded subject appearance in $f_{GAN}$.
 % and therefore is able to provide a more accurate supervision signal for the relighting network. 
 
\subsection{Training Objective} \label{obj}
To ensure both realness and fidelity, we use the following terms in our training objective function: 

\noindent\textbf{Reconstruction Loss} $\m{L}_{R_{norm}}$, $\m{L}_{R_{alb}}$ and $\m{L}_{R_{relit}}$: The standard $L1$ distance between the predicted normal, albedo, relit image and their ground-truth counterparts. 

\noindent\textbf{Perceptual Loss} $\m{L}_{P_{alb}}$ and $\m{L}_{P_{relit}}$: The layer-wise feature difference between the predicted albedo/relit portrait and the ground truth albedo/relit portrait extracted by a pretrained VGG~\cite{simonyan2014very} to increase perceptual quality. 

\noindent\textbf{Relighting Adversarial Loss} $\m{L}_{A_{relit}}$: The adversarial loss between the predicted relit portrait and ground truth to encourage the generation of high-frequency lighting details. We adopt a PatchGAN discriminator with spectral normalization~\cite{yu2019free} to compute this loss. 

For delighting, We jointly train the albedo and normal net by optimizing:
\begin{align}
    \m{L}_{delight} = \m{L}_{R_{norm}} + \m{L}_{R_{alb}} +\m{L}_{P_{alb}}
\end{align}
For the relighting stage, the totally 
loss can be written as:
\begin{align}
\m{L}_{relight} =\m{L}_{R_{relit}} + \m{L}_{P_{relit}} + \m{L}_{A_{relit}}
\end{align}

\section{Data and Implementation Details}~\label{data}
\vspace{-8mm}
\paragraph{Data Preparation.} Following~\cite{pandey2021total,sun2019single,mei2023lightpainter}, we use light stage captures ~\cite{deng2019accurate} to render high quality datasets. Our light stage is structurally similar to~\cite{mei2023lightpainter,sun2019single}, which consists of 160 programmable LED lights and 4 frontal-view cameras.
The light stage captures contain 69 subjects photographed with different poses, expressions and accessories, resulting in total 931 OLAT sequences. We reserve $15$ subjects with different genders and races for testing. Following~\cite{pandey2021total}, we use the image captured with flat omnidirectional lighting (\ie all lights turned on) as ground-truth albedo and use photometric stereo~\cite{wenger2005performance} to calculate normals.

To create a high-quality training dataset, we collect 671 real environment maps from PloyHeaven~\cite{ployhaven}. We randomly select a subset of 500 environment maps for training and use the rest for testing. We augment lighting by randomly rotating the environment map and further include the original OLAT images into the training dataset, which results in a total of 520K images. We randomly pair testing OLAT sequence and environment maps to create test sets. More details can be found in the supplemental file. 
\vspace{-3mm}
\paragraph{Training Details.}

For training, we 
group 8 faces crops of size $512\times 512$ as one batch. All networks are optimized with Adam~\cite{KingBa15} with a learning rate of $1e-5$. 
% We stop training after 10 epochs. 
We joint train the albedo and normal net for 5 epochs, and train the relighting net for 10 epochs.
The volume rendering resolution is set to $64\times 64$. We adopt pre-trained EG3D and finetune their super-resolution network on our dataset. We freeze the tri-plane generator $G_{tri}$ throughout the training. 
The proposed model is implemented using PyTorch and trained on 8 A100 GPUs. More details can be found in the supplement. 

%\section{Experiments}
\begin{figure*}[t!]
	\centering
	\includegraphics[clip, trim=0cm 8.5cm 12.1cm 0cm, width=1\textwidth]{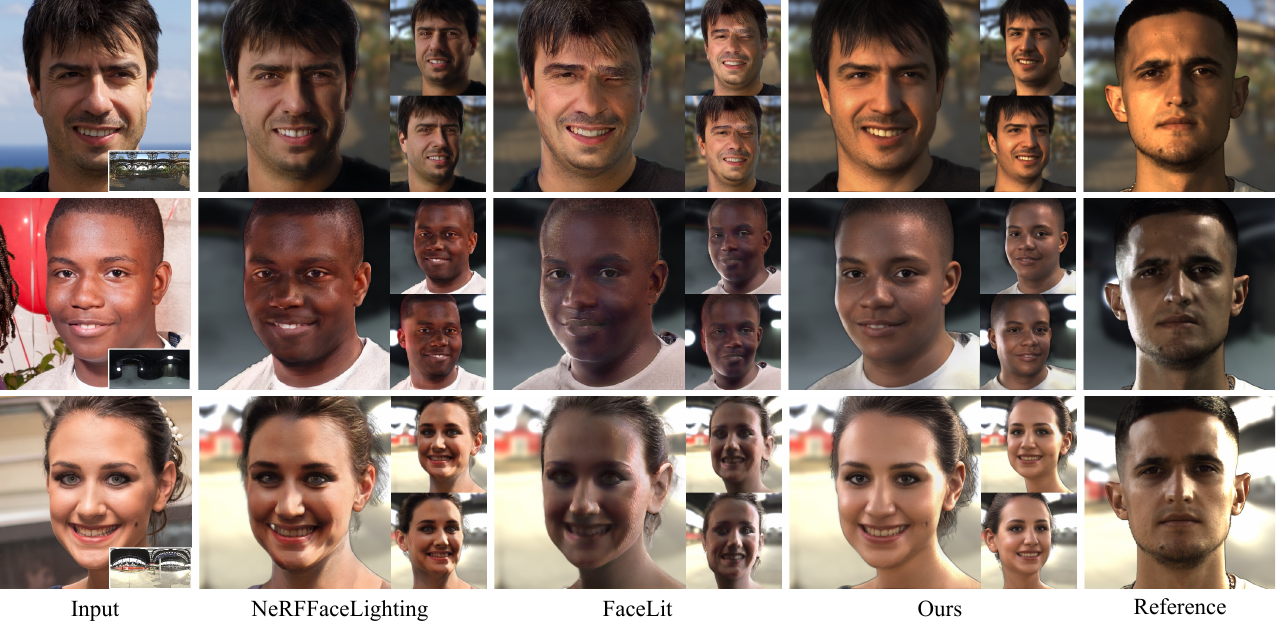}
\vskip-8pt	
\caption{\textbf{Visual comparisons of free-view relighting on in-the-wild portraits.}
Environment maps are shown as insets. We provide a reference image (last column) rendered using OLAT subject with the target lighting as guidance for comparison.
}\label{fig:wild-3d}
\vspace{-5mm}
\end{figure*}
\begin{figure}[t!]
	\centering
	\includegraphics[clip, trim=0.25cm 5cm 12cm 0cm, width=\columnwidth]{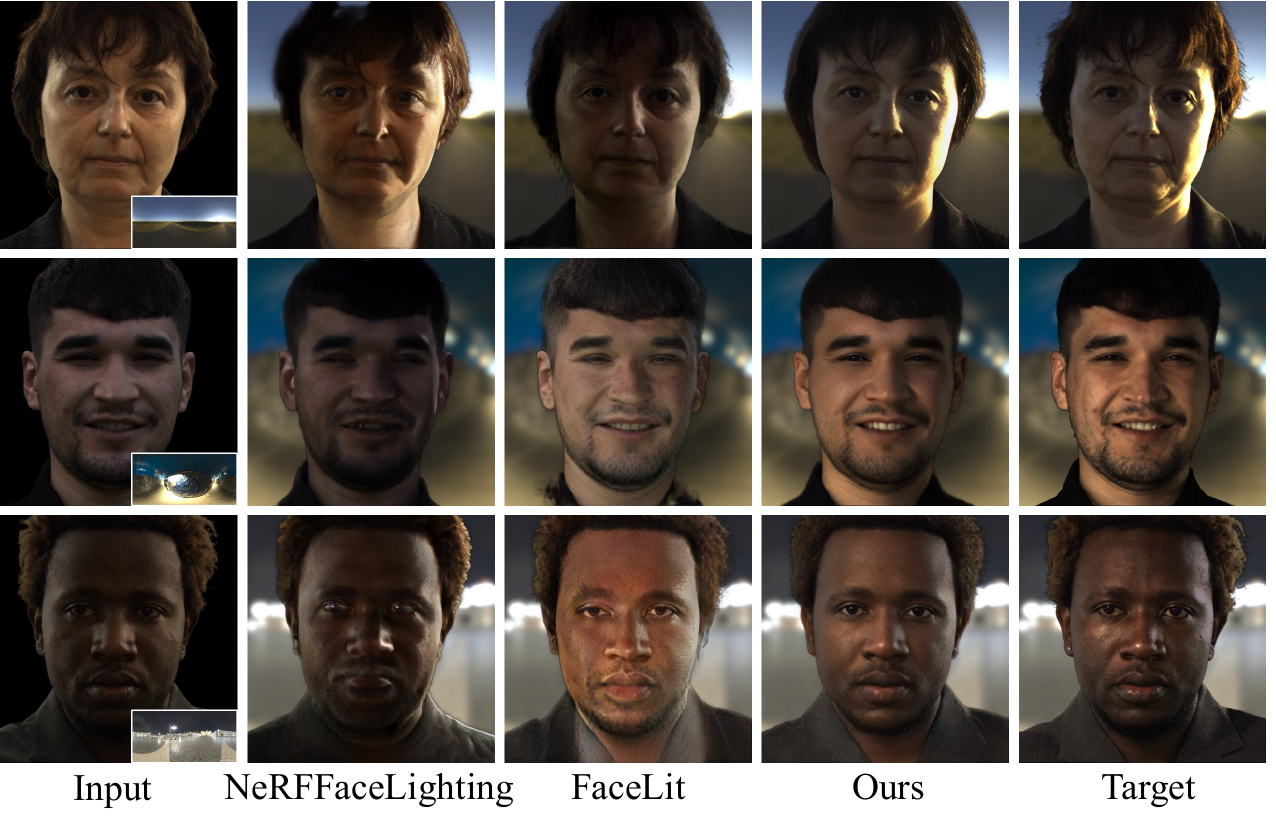}
\vskip-8pt	\caption{\textbf{Visual comparisons against free-view relighting methods on test set.} Our method produces more faithful target lighting effects.}\label{fig:3d-val}
\vspace{-5mm}
\end{figure}
\section{Experiments}\label{experiment}In this section, we demonstrate the relighting quality and control capability of \agoodname through extensive experiments. More results can be found in the supplemental file.
\vspace{-5mm}
\paragraph{Evaluation Metrics.} We follow~\cite{mei2023lightpainter} and report LPIPS~\cite{zhang2018unreasonable} and NIQE~\cite{zhang2015feature} for perceptual quality and PSNR and SSIM~\cite{wang2004image} for fidelity. We also report identity metrics Deg (cosine similarity between LightCNN~\cite{wu2018light} features) to measure identity preservation. All metrics are calculated on the foreground subject with pre-computed masks using~\cite{MODNet}.

\subsection{Comparisons with State-of-the-art Methods}
We compare  \agoodname with state-of-the-art free-view relighting methods FaceLit~\cite{ranjan2023facelit}, and NeRFFaceLighting~\cite{jiang2023nerffacelighting}. In addition, as \agoodname also supports conventional 2D portrait relighting by rendering to the input pose, we further show its effectiveness by comparing it with the state-of-the-art 2D relighting method Total Relighting~\cite{pandey2021total}(TR). For both experiments, we perform qualitative evaluations on in-the-wild images, and quantitative \& qualitative evaluations on our test set. 

\begin{table}[t]\label{tab:user}
       
        \small
        %\scriptsize
        \centering
        \caption{
            % Quantitative evaluation on validation dataset.
            Quantitative evaluations against free-view relighting methods.
            }
         \vspace{-2mm}
        \tabcolsep=0.06cm
        {
                %\hspace{-0.5cm}
                \begin{tabular}{l|cc|c|cc}
                        \hline
                        Methods       & LPIPS$\downarrow$   &NIQE$\downarrow$    & Deg$\uparrow$   & PSNR$\uparrow$  & SSIM$\uparrow$  \\ \hline \hline
                              NeRFFaceLighting~\cite{jiang2023nerffacelighting}    &0.2179 &8.132 &0.6953 &19.40  &0.7214 \\ 
                        FaceLit~\cite{ranjan2023facelit}            &0.1542   &8.017 &0.7821 & 20.86 &0.7281  \\ \hline
                        \textbf{Ours} & \textbf{0.0917}  & \textbf{5.273}& \textbf{0.8978}&\textbf{27.35}  &\textbf{0.8511} \\ \hline
                        
        \end{tabular}}\label{tab:3d-gan}
        \vspace{-3mm}
\end{table}

\begin{table}[t]\label{tab:user}
       
        \small
        %\scriptsize
        \centering
        % \caption{Quantitative evaluation on validation dataset for 2D protrait Relighting.}
        \caption{
        Quantitative evaluations against 2D relighting methods.
        }
         \vspace{-2mm}
        \tabcolsep=0.22cm
        \resizebox{\columnwidth}{!}{
                %\hspace{-0.5cm}
                \begin{tabular}{l|cc|c|cc}
                        \hline
                        Methods      & LPIPS$\downarrow$   &NIQE$\downarrow$    & Deg$\uparrow$   & PSNR$\uparrow$  & SSIM$\uparrow$  \\ \hline \hline
                              TR~\cite{pandey2021total}    &0.1997 &7.012 &0.7638 &19.58  &0.6515 \\ 
                        \textbf{Ours} & \textbf{0.0981}  & \textbf{6.195}& \textbf{0.8793}&\textbf{26.17}  &\textbf{0.8544} \\ \hline
                        
        \end{tabular}}\label{tab:2d}
        \vspace{-5mm}
\end{table}

\subsubsection{Free-view Portrait Relighting.} 
To evaluate the performance on both view-synthesis and relighting, we compare our method with two state-of-the-art works: FaceLit~\cite{ranjan2023facelit} and NeRFFaceLighting~\cite{jiang2023nerffacelighting}. For both methods, we follow their papers to estimate Spherical Harmonics coefficients from a reference image as target lighting (by adopting lighting estimator from \cite{feng2021learning} and \cite{zhou2019deep} respectively). 
The reference image is rendered with environment maps and OLAT data~\cite{debevec2000acquiring} throughout our experiments.
To perform relighting on an input image, we use \cite{xie2023high} to perform inversion for FaceLit~\cite{ranjan2023facelit} as it does not have a native inversion module. For a fair comparison, we fix the head pose to frontal in our system throughout the evaluation to accommodate the setting in \cite{ranjan2023facelit} and \cite{jiang2023nerffacelighting}. 
\vspace{-3mm}
\begin{figure}[t!]
	\centering
	\includegraphics[clip, trim=1.1cm 2cm 9cm 0cm, width=0.5\textwidth]{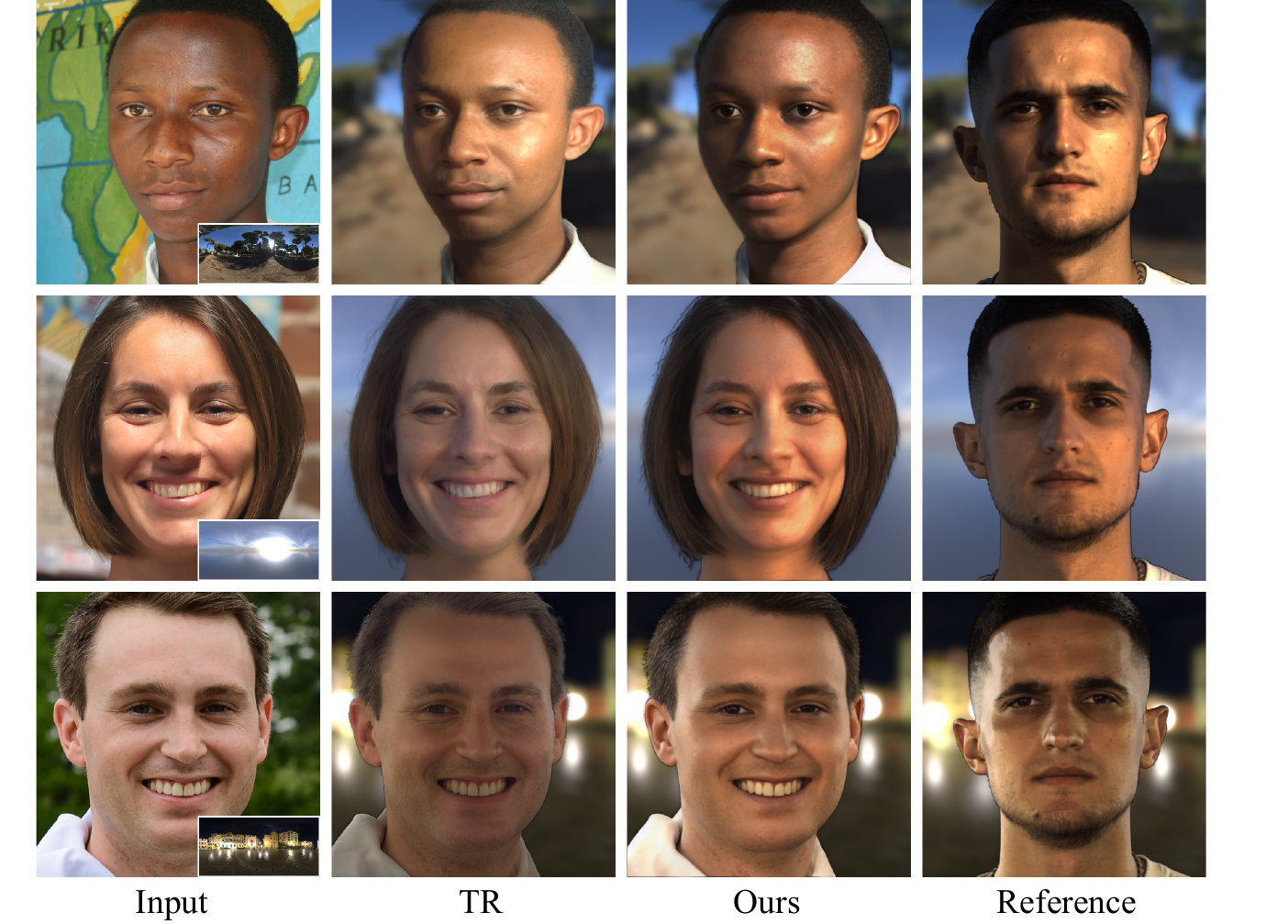}
\vskip-8pt	\caption{\textbf{Visual comparisons of 2D portrait relighting on in-the-wild images.} We provide a reference image rendered using OLAT subject and target lighting as guidance for comparsion. }\label{fig:2d-wild}
\vspace{-4mm}
\end{figure}
\begin{figure}[t!]
	\centering
	\includegraphics[clip, trim=0cm 2.cm 11cm 0cm, width=0.5\textwidth]{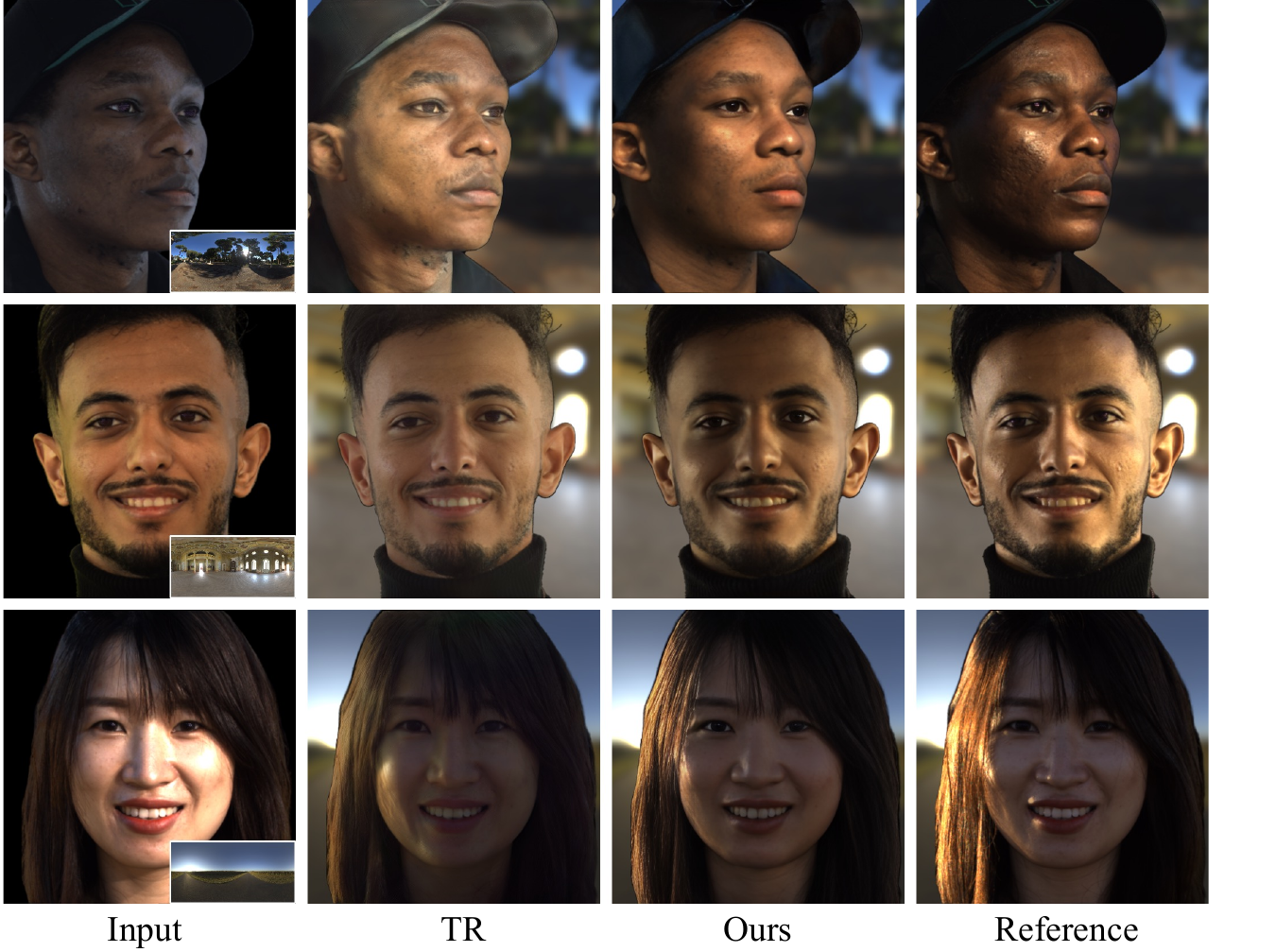}
\vskip-5pt	\caption{\textbf{Visual comparisons of 2D portrait relighting on test set}.  Our method produces on-par or better results than Total Relighting~\cite{pandey2021total} (TR).}\label{fig:val-2d}
\vspace{-3mm}
\end{figure}

\vspace{-1mm}
\paragraph{Qualitative results on in-the-wild images.} We demonstrate the generalization capability of \agoodname by performing relighting on in-the-wild portrait images, as shown in Figure~\ref{fig:wild-3d}. 
Since there is no ground truth, we provide a reference image rendered with the target environment map in the last column for perceptual comparison. 
\agoodname generates the most convincing results, among all methods, with realistic shading and specularity. It is also robust to various lighting conditions, producing lighting effects that closely matches the reference images. In contrast, the results from NeRFFaceLighting and FaceLit contain visual artifacts and cannot fully reproduce the lighting effects. 
\vspace{-5mm}
%\paragraph{Results on the OLAT Test Set.} 
\paragraph{Quantitative and qualitative results on OLAT test set.} 
For quantitative evaluation, we create a test dataset of 235 images using OLAT data (frontal head pose) and sampled environment maps. 
Quantitative results are reported in Table~\ref{tab:3d-gan}. Our method consistently achieves the best perceptual quality, fidelity, and identity preservation quality. 
We report qualitative results in Figure~\ref{fig:3d-val}. Given an input and a target lighting, our results match the target image most closely with faithful lighting effect and superior visual quality.

\begin{figure}[t!]
	\centering
	\includegraphics[clip, trim=0.2cm 10.6cm 17.5cm 0.1cm, width=\columnwidth]{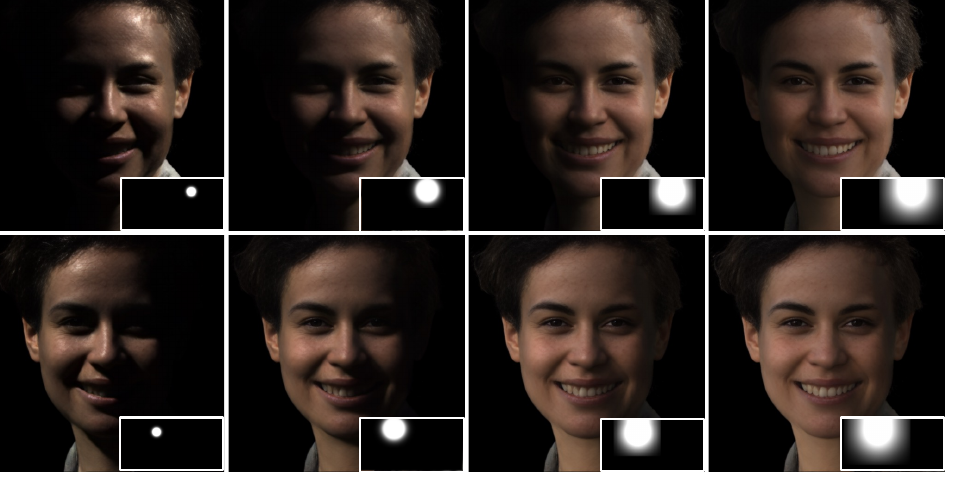}
\vskip-11pt	\caption{\textbf{Illustration of lighting control using a single light source.} Our method produces realistic cast shadows and specular reflection. Important applications such as shadow diffusion
(softening) can be easily achieved with \agoodname.}\label{fig:shadow_diff}
\vspace{-5mm}
\end{figure}

\begin{figure}[t!]
	\centering
	\includegraphics[clip, trim=0.1cm 3.5cm 18.4cm 0.1cm, width=1\columnwidth]{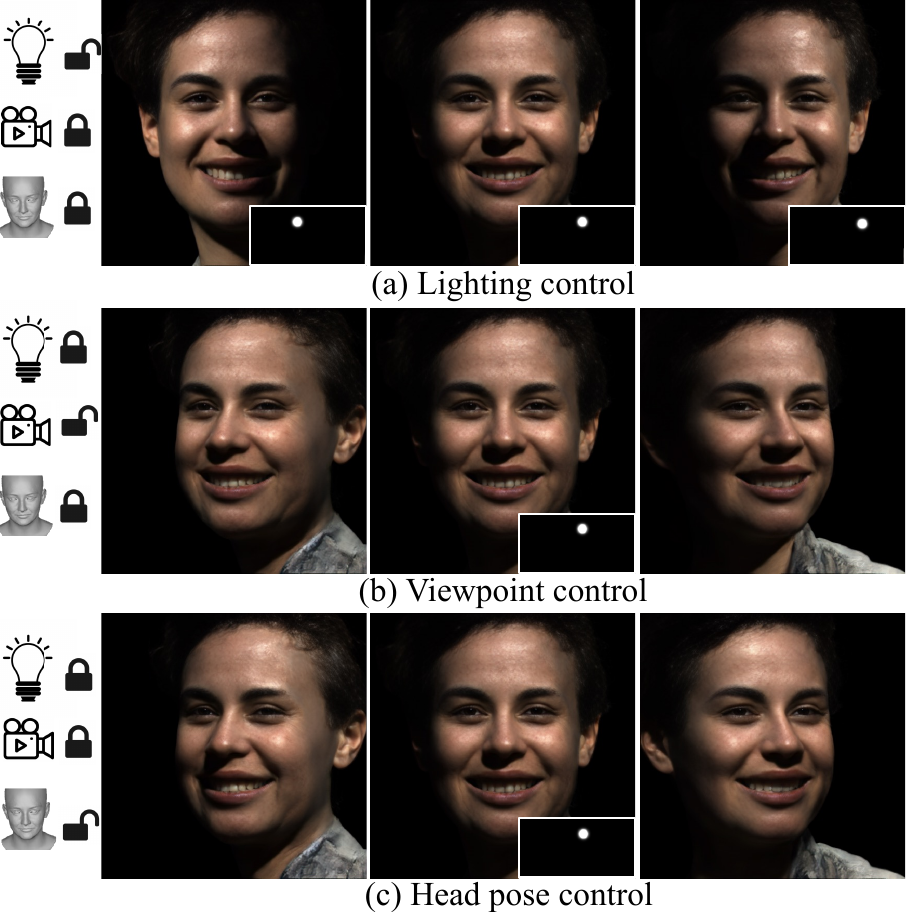}
\vskip-6pt	\caption{\textbf{Illustration of controllability using a single light source.} Our method produces consistent shadows and specular highlights with respect to a rotating (a) light source, (b) camera pose and (c) head pose.}\label{fig:shadow_syth}
\vskip-8pt	
\end{figure}

\begin{comment}
    \begin{figure}[t!]
	\centering
	\includegraphics[clip, trim=0.2cm 7.5cm 18cm 0.1cm, width=0.49\textwidth]{figures/specular_pose.pdf}
\vskip-5pt	\caption{\textbf{viewpoint based and head-pose based lighting control}}\label{fig:head-pose}
\label{app_headpose}
\vspace{-2mm}
\end{figure}
\end{comment}

\subsubsection{2D portrait relighting.} 
We compare \agoodname with the state-of-the-art relighting method Total Relighting~\cite{pandey2021total} (TR) on 2D portrait relighting. Total relighting is an image-to-image based system trained with OLAT data, and therefore does not allow view synthesis.

\vspace{-5mm}
\paragraph{Qualitative results on in-the-wild images.} We report visual comparison results in Figure~\ref{fig:2d-wild}. Our method produces photo-realistic diffuse and high-frequency specular reflections and robustly handles diverse subjects and illumination. Compared to TR~\cite{pandey2021total}, our method produces lighting effects that are more similar to the corresponding reference images.  
\vspace{-5mm}
\paragraph{Quantitative and qualitative results on OLAT test set.} We create a test set of 940 images using OLAT data. From Figure~\ref{fig:val-2d},
we can see that our method produces more faithful renderings than TR. 
% faithfully renders the target illumination compared to TR. 
As shown in Table~\ref{tab:2d}, our method also yields better quantitative results. 

\subsection{Controllability} We demonstrate the flexible controllability of \agoodname by illuminating an in-the-wild portrait photo~\footnote{The original image can be found in the supplemental file.} under a single directional light source. Using just one light source can effectively highlight intricate lighting effects such as specular highlights and cast shadows and showcase sophisticated light transport phenomena,
which are crucial components for photorealism. Please refer to the supplemental video for more examples.

\vspace{-3mm}
\paragraph{Lighting Control.} ~\agoodname can generate realistic cast shadows and specular highlights from a single light source, as shown in the first column of Figure~\ref{fig:shadow_diff}. By softening the light source through blurring the environment map, the network can generate realistic shadow diffusion (softening) effects on the face.  
As shown in Figure~\ref{fig:shadow_syth}-(a), \agoodname produces consistent moving shadows and specular reflections with a rotating light source. 
\vspace{-3mm}
\paragraph{Viewpoint Control.} In addition to the view synthesis results shown in Figure~\ref{tab:3d-gan}, we further demonstrate our view control capability through the consistent lighting effect in Figure~\ref{fig:shadow_syth}-(b): we can generate consistent cast shadows (\eg under noise) and specular highlights (\eg on the forehead and cheek) across different views.
\vspace{-3mm}
\paragraph{Pose Control. } 
% We demonstrate a unique head pose control 
We demonstrate our unique support for head pose control in Figure~\ref{fig:shadow_syth}-(c). By comparing it with images of the same column in Figure~\ref{fig:shadow_syth}-(b), one can see that \agoodname synthesizes realistic moving highlights (\eg on the forehead) and cast shadows (\eg under the noise) with head rotation.

\subsection{Ablation Study}
We conduct ablation study on two proposed data-rendering methods: (1) Multi-view GAN Inversion and (2) Shading Transfer (ST). We report quantitative results in Table~\ref{tab:arch}. We can conclude that (1) multi-view inversion significantly improves the visual quality of the result compared to vanilla single-view inversion during the training phase; (2) shading transfer helps to align the appearance of the target image with that encoded in GAN features and further improve the performance of \agoodname. More details and qualitative results can be found in the supplemental material.
\begin{table}[t]
       
        \small
        %\scriptsize
        \centering
        \caption{Ablation study on proposed data-rendering strategies.}
         \vspace{-3mm}
        \tabcolsep=0.13cm
        \resizebox{\columnwidth}{!}{
                %\hspace{-0.5cm}
                \begin{tabular}{l|cc|c|cc}
                        \hline
                        Methods       & LPIPS$\downarrow$   &NIQE$\downarrow$    & Deg.$\uparrow$   & PSNR$\uparrow$  & SSIM$\uparrow$  \\ \hline \hline
                        
                        Single-view          &0.1443  &7.381 &0.8142 &23.57 &0.7845 \\  
                        w/o ST          &0.1382   &7.143 &0.8387 &24.03 &0.7840  \\ \hline 
                        \textbf{\agoodname} & \textbf{0.0997}  & \textbf{6.330}& \textbf{0.8735}&\textbf{25.89}  &\textbf{0.8479} \\ \hline
                        
        \end{tabular}
        }
        \label{tab:arch}
        \vspace{-6mm}
\end{table}

\section{Conclusion}
In this work, we propose \agoodname, a novel volumetric relighting method that is capable of synthesizing renderings under novel viewpoints and novel lighting from 
a single input image. 
\agoodname leverages the 3D representations from the pretrained 3D-aware GAN (EG3D) and predicts a lighting and pose-aware 3D representation (tri-plane features), from which a portrait image with controllable lighting, head pose, and viewpoint can be produced using volume rendering.
We demonstrate the high controllability and state-of-the-art relighting quality of \agoodname through extensive experiments. Discussion of limitation can be found in the supplemental file.

\noindent\textbf{Acknowledgments}
This work was supported by NSF
CAREER award 2045489. We thank Chaowei Company for the support of light stage data.

{
    \small
    \bibliographystyle{ieeenat_fullname}
    \bibliography{main}
}

\end{document}